\documentclass[conference]{IEEEtran}
\IEEEoverridecommandlockouts
\usepackage{cite}
\usepackage{amsmath,amssymb,amsfonts}
\usepackage{graphicx}
\usepackage{textcomp}
\usepackage{xcolor}
\usepackage{multirow}
\usepackage{multicol}
\usepackage{graphicx}
\usepackage{booktabs}
\usepackage{algorithm}
\usepackage{colortbl}
\usepackage{algpseudocode}

\def\BibTeX{{\rm B\kern-.05em{\sc i\kern-.025em b}\kern-.08em
    T\kern-.1667em\lower.7ex\hbox{E}\kern-.125emX}}
\begin{document}

\title{PatchINR: Patch-Based Implicit Neural Representations for Efficient and Scalable Inference}
\author{
    \IEEEauthorblockN{Jiachen Ren$^{\dagger}$, Wenyong Zhou$^{\dagger}$, Taiqiang Wu, Yuxin Cheng, Xincheng Feng, Zhengwu Liu$^*$, Ngai Wong$^*$}
    \IEEEauthorblockA{
        The Department of Electrical and Computer Engineering, The University of Hong Kong, Hong Kong.
    }
    \IEEEauthorblockA{
        \textsuperscript{\dag}: Equal contributions. $^*$: Corresponding authors.
    }
}
\maketitle
\begin{abstract}
Implicit Neural Representation (INR) provides an effective approach for continuous signal modeling, but classical per-pixel inference results in quadratic growth in inference count, leading to dramatically increased computational costs in high-resolution application scenarios. To address this issue, we propose a patch-based approach that treats non-overlapping patches as fundamental processing units and predicts entire pixel patches in a single forward pass, significantly reducing the number of inference queries required. To validate the effectiveness of our approach, we propose a hardware acceleration architecture on the Field Programmable Gate Array (FPGA) platform for the INR model, which features a configurable pipeline and supports dual-precision computation. Our patch-based INR achieves comparable reconstruction quality to pixel-level INR (34.97 dB PSNR with $2 \times 2$ patches) while reducing inference latency by 75\% with only 0.6\% parameter overhead.
\end{abstract}

\section{Introduction}
\label{sec:introduction}

Implicit Neural Representations (INRs) have emerged as a powerful paradigm for representing continuous signals, offering compact and continuous representations of images, 3D shapes, and other data modalities~\cite{deepsdf, metasdf}. By mapping spatial coordinates to signal values through neural networks, INRs provide a memory-efficient alternative to discrete grid-based methods while enabling high-fidelity signal reconstructions~\cite{chang, int}. Their inherent resolution independence and the ability to query arbitrary coordinates make them highly ideal for advanced applications such as neural rendering, 3D reconstruction, image super-resolution, and dynamic scene representation~\cite{gradient, tancik2020fourier}. However, deploying these computationally intensive continuous representations on resource-constrained edge devices or real-time systems poses a formidable challenge, necessitating highly optimized inference frameworks and dedicated hardware acceleration.

\begin{figure}[!t]
\centering
\includegraphics[width=1.0\columnwidth]
{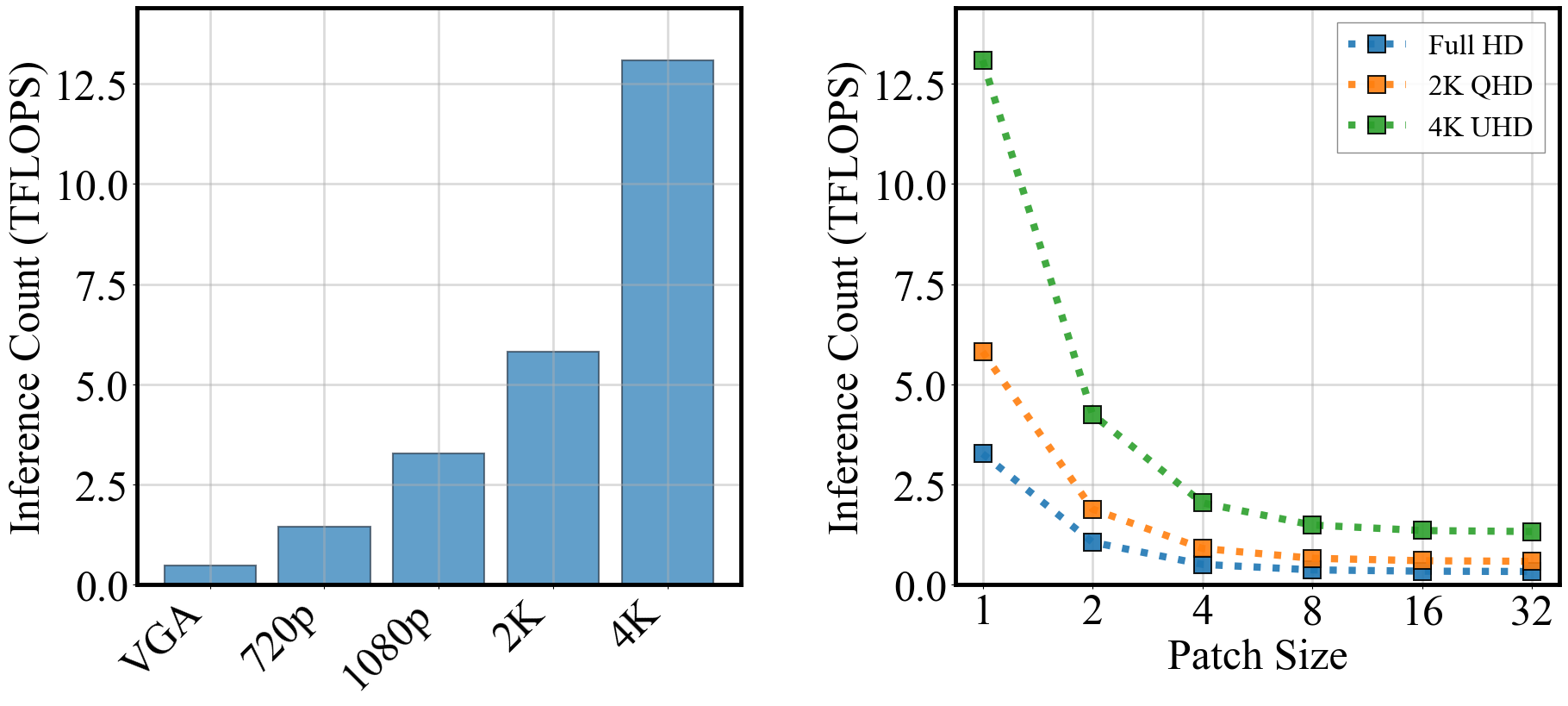}
\vspace{-0.3cm}
\caption{(Left) Inference count grows quadratically as the resolution increases for the image. (Right) Our patch-based method significantly reduces the inference count.}
\label{fig:motivation_data}
\end{figure}

Despite their representational advantages, traditional INRs inherently rely on a pixel-wise querying paradigm, where the network must evaluate each spatial coordinate independently to reconstruct the target signal. From a hardware perspective, this sequential and fine-grained evaluation paradigm prevents the efficient utilization of modern parallel computing architectures, which are fundamentally optimized for bulk data processing and spatial parallelism~\cite{minr}. Consequently, this pixel-level processing approach requires millions of redundant coordinate fetches and independent forward passes to generate a single high-resolution image, rendering INRs practically infeasible for latency-sensitive and real-time applications~\cite{inr_quant}. As illustrated in Fig.~\ref{fig:motivation_data}, the total inference workload grows quadratically with the target spatial resolution~\cite{dhq}. This quadratic complexity arises directly from the inherent inefficiency of the pixel-by-pixel evaluation strategy: each isolated pixel query incurs an independent forward pass and redundant scalar computations. To address this systemic computational bottleneck, we propose a paradigm shift towards a patch-based optimization strategy that processes multiple pixels simultaneously within spatial regions, thereby drastically reducing the total number of forward passes and enabling efficient batch processing.

Previous efforts to mitigate INR inefficiency have predominantly focused on algorithmic modifications, such as lightweight network designs or grid-based hybrid representations~\cite{coordx,asmr}. However, these software-centric optimizations often fail to address the underlying hardware unfriendliness of the querying mechanism itself. To bridge this gap, we introduce a hardware-algorithm co-designed framework: a novel patch-based INR that processes spatial information by predicting entire $n \times n$ patches of pixel values in a single unified forward pass, rather than querying individual pixel coordinates. By shifting the fundamental computational granularity from isolated pixels to non-overlapping patches, our method reduces the total number of inference queries by a factor of $n^2$. This transformation effectively amortizes the coordinate fetching overhead and significantly enhances data reuse within on-chip buffers. Crucially, our approach preserves the continuous, resolution-independent properties of traditional INRs while exposing massive spatial parallelism that maps seamlessly onto hardware accelerators, perfectly aligning with deeply pipelined architectures and resource utilization strategies. In summary, the contributions of this paper are as follows:

\begin{itemize}
\item We propose a novel patch-based INR framework that redefines the fundamental querying unit from individual pixels to non-overlapping spatial patches. This algorithmic innovation achieves a significant reduction in total inference costs and computational overhead, while strictly maintaining reconstruction fidelity comparable to classical pixel-wise INR approaches.
\item We design and implement a highly optimized Field Programmable Gate Array (FPGA) acceleration architecture specifically tailored for the patch-based SIREN model. The proposed accelerator features a deeply configurable pipeline, robust support for dual-precision computation, and accommodates parameterized patch dimensions to maximize hardware utilization.
\item Extensive experimental evaluations demonstrate that our co-designed method achieves an optimal trade-off between inference efficiency and model parameter count. Specifically, utilizing a $2 \times 2$ patch size reduces the overall inference latency by 75\% and achieves a high reconstruction quality of 34.97 dB PSNR, while incurring a negligible parameter overhead of merely 0.6\%.
\end{itemize}

\section{Related Work}
\label{sec:related}
\subsection{Implicit Neural Representation}
INRs have gained significant traction as a compact and flexible approach for representing continuous signals. By parameterizing signals as neural networks that map spatial coordinates to signal values, INRs eliminate the need for discrete grid-based storage~\cite{diner,cicero}. Traditional INRs rely on pixel-wise querying, which becomes computationally prohibitive for high-resolution or real-time applications. To address the inefficiency of pixel-wise querying, various strategies have been proposed. Hierarchical representations and hybrid approaches combining grids and neural networks~\cite{coordx} have been explored to accelerate inference. Additionally, techniques such as frequency encoding~\cite{tancik2020fourier} and efficient architectural designs~\cite{shacira} aim to speed up computation while preserving reconstruction quality. While these methods provide moderate improvements, they remain constrained by the underlying pixel-wise querying paradigm, making them less suitable for real-time or large-scale applications.

\subsection{Hardware Acceleration for INR}
FPGA-based inference acceleration has demonstrated compelling advantages in energy efficiency, ultra-low latency for real-time processing, and parallel computing capabilities for high-throughput workloads, making it particularly suitable for deploying deep learning models on edge devices. The deployment of INRs on hardware accelerators has garnered significant attention due to the growing demand for real-time processing and high-fidelity rendering in downstream applications. However, the inherent sequential nature of traditional pixel-wise querying introduces severe memory access bottlenecks, preventing INRs from fully utilizing the massive spatial parallel processing capabilities inherent in modern hardware architectures. Recent efforts to optimize INRs for hardware deployment include model quantization to reduce memory footprint~\cite{nerf_quant}, energy-efficient dataflow designs~\cite{inreetcasi}, and memory-efficient architectures aimed at alleviating bandwidth limitations~\cite{asmr}. Despite these commendable advancements, the lack of a fundamentally batch-processing-friendly algorithmic framework has severely limited the scalability, throughput, and practical deployment of INRs, particularly for high-resolution tasks in resource-constrained environments.
\section{Methodology}
\label{sec:methodology}
\subsection{Patch-Based INR}
Traditional INRs map spatial coordinates \( x \in \mathbb{R}^d \) to their corresponding signal values \( y \in \mathbb{R}^c \) using a neural network \( f_\theta \). This process can be expressed as:
\begin{equation}
y = f_\theta(x)
\end{equation}
where \( \theta \) represents the network parameters. This formulation enables continuous signal representation and resolution independence, as any coordinate in the input domain can be queried to generate the corresponding signal value. For example, an image \( I \in \mathbb{R}^{H \times W \times c} \) can be reconstructed by querying all spatial coordinates \( x_{i,j} \), with the signal at each coordinate defined as:
\begin{equation}
y_{i,j} = f_\theta\left(\frac{i}{H}, \frac{j}{W}\right), \quad \forall \, i, j \in \{1, \dots, H\} \times \{1, \dots, W\}.
\end{equation}

\begin{figure}[!t]
\centering
\includegraphics[width=1.0\columnwidth]{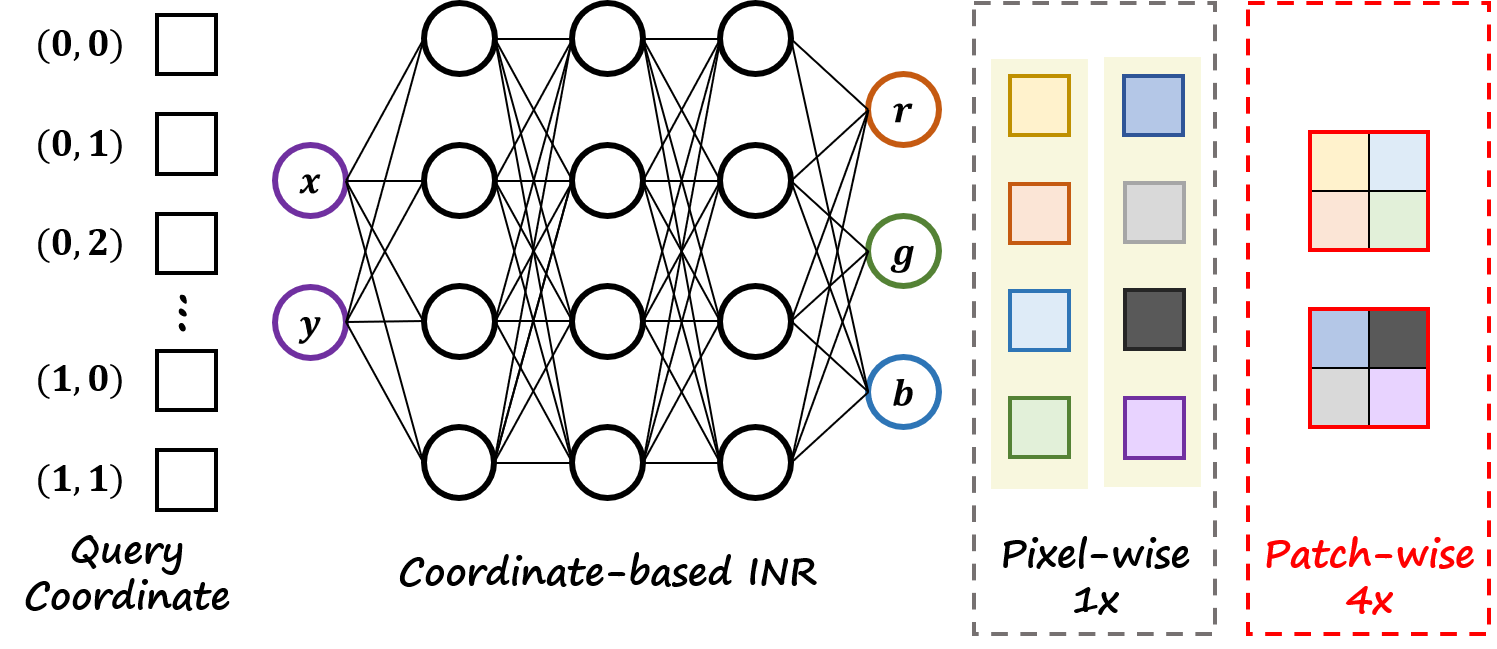}
\vspace{-0.3cm}
\caption{Unlike classical INRs that transform each query coordinate to its corresponding pixel, our proposed patch-based INR significantly accelerates inference by generating multiple pixels simultaneously through patch-wise representation.}
\label{fig:motivation}
\end{figure}

\begin{figure*}[!t]
\centering
\includegraphics[width=1.0\textwidth]{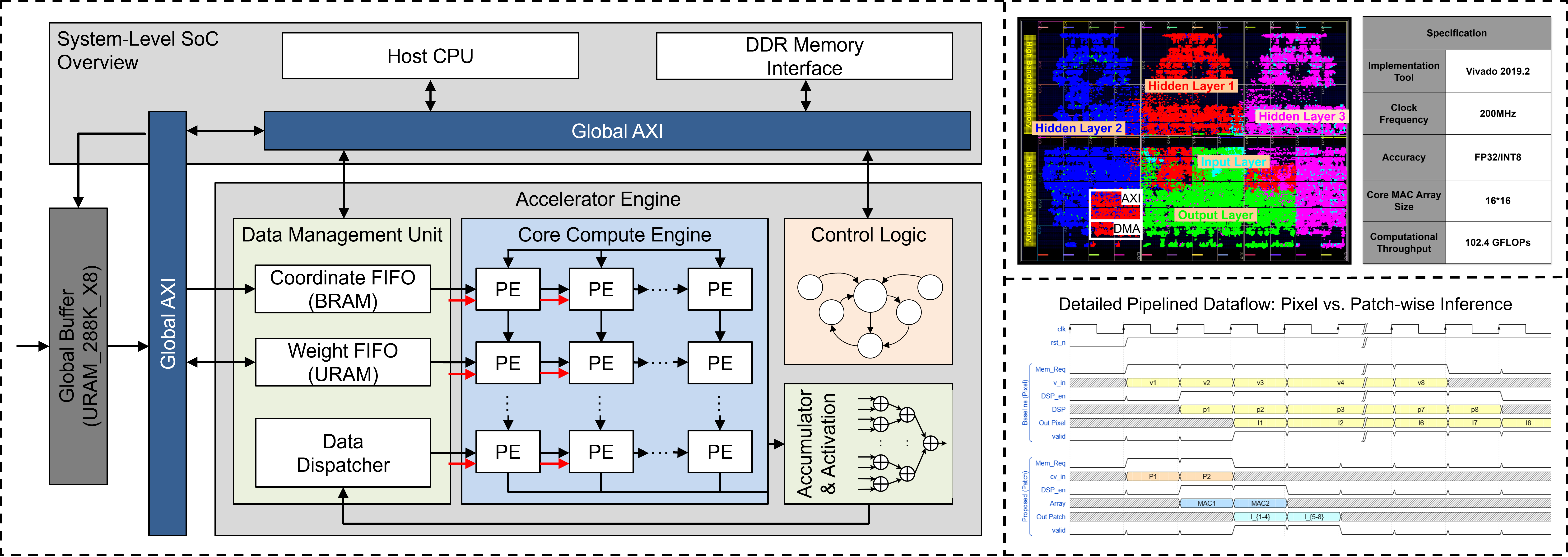}
\caption{Hardware implementation of the PatchINR accelerator. The design features a system-level SoC architecture with a parallel compute engine (left), an optimized physical FPGA layout for neural network execution (top right), and a pipelined dataflow timeline demonstrating significant latency reduction for patch-based inference (bottom right).}
\label{fig:hardware_system}
\end{figure*}
Instead of querying single coordinates, we treat non-overlapping regions (patches) of size \( P \times P \) as a single entity and define a new grid corresponding to these patches, as shown in Fig.~\ref{fig:motivation}. For example, a patch size of \( P = 2 \) would correspond to \( 2 \times 2 \) pixel regions, and the input signal \( I \in \mathbb{R}^{H \times W \times c} \) is divided into non-overlapping patches, where \( P \) is the patch dimension hyperparameter. The new grid is defined by coordinates \( x_p \in \mathbb{R}^d \), with a total of:
\begin{equation}
N = \frac{H \cdot W}{P^2}
\end{equation}
patches, where \( H \) and \( W \) are the height and width of the original signal. Each patch corresponds to a unique coordinate in the new grid. 

To represent the signal in this patch-based framework, we reformulate the INR function \( f_\theta \). For a given patch coordinate \( x_p \), the INR predicts the pixel values for the entire \( P \times P \) patch, expressed as:
\begin{equation}
Y_p = f_\theta(x_p),
\end{equation}
where \( Y_p \in \mathbb{R}^{P \times P \times c} \) represents the predicted pixel values for the corresponding patch. The INR model \( f_\theta \) is still queried using continuous spatial coordinates, but now operates on the new, coarser grid defined by the patches.

This patch-based formulation greatly reduces the number of queries required to reconstruct the signal. Instead of querying \( HW \) individual coordinates, the number of queries is reduced to \( N \). The computational complexity of inference is thus reduced from $\mathcal{O}(HW \cdot T)$ to $\mathcal{O}(N \cdot T)$, where \( T \) is the computation time per query. For larger patch sizes \( P \), \( N \) becomes significantly smaller than \( HW \).

\subsection{Hardware Implementation}
While algorithmic optimizations reduce theoretical computational complexity, traditional sequential coordinate evaluation remains severely restricted by irregular memory access patterns. To validate the proposed patch-based framework and achieve highly efficient deployment, we developed a customized hardware accelerator architecture as illustrated in Fig.~\ref{fig:hardware_system}. The overall design adopts a system-on-chip integration paradigm, connecting a host processor, off-chip Double Data Rate (DDR) memory, and the custom accelerator engine through a global Advanced eXtensible Interface (AXI) interconnect. To mitigate the severe bandwidth bottlenecks associated with fetching massive neural network weights and spatial coordinates, a high-capacity global buffer instantiated with Ultra Random Access Memory (URAM) is introduced to cache frequently accessed operands.

Within the accelerator engine, a dedicated data management unit orchestrates the on-chip dataflow. This unit employs Block Random Access Memory (BRAM) for the coordinate queue and URAM for the weight queue. A dedicated data dispatcher subsequently routes these synchronized data streams directly into the core compute engine to ensure a continuous supply of operands.

The core compute engine serves as the computational backbone of the accelerator, featuring a highly parallel two-dimensional array of processing elements. Each processing element is deeply pipelined and maps the matrix multiplication workloads onto specialized Digital Signal Processing (DSP) slices. These custom slices are meticulously configured with input multiplexers, hardware multipliers, and local registers to execute the dense multiply-accumulate operations required by the hidden layers. By leveraging this extensive spatial parallelism, the compute engine processes multiple pixels within a spatial patch concurrently. This hardware mapping effectively translates the algorithmic advantages of our batch processing formulation into tangible throughput improvements.

To ensure data integrity, a centralized Finite State Machine (FSM) synchronizes the pipelined dataflow and manages execution states. The processing element outputs are subsequently routed to a unified accumulator for vector aggregation and nonlinear activation. Finally, the reconstructed patches are written back to the global memory hierarchy. This co-design effectively mitigates the memory wall and maximizes on-chip resource utilization.
\section{Experiments}
\label{sec:experiments}
\subsection{Experiment Setup}
To comprehensively evaluate the proposed patch-based framework, we establish a rigorous software-hardware co-evaluation environment. On the algorithm side, we select three representative state-of-the-art INR architectures, SIREN~\cite{siren}, WIRE~\cite{wire}, and FINER~\cite{finer} as our baselines. These models are evaluated on standard high-fidelity image reconstruction benchmarks, specifically the Kodak and DIV2K datasets~\cite{siren, gaussian}. The models are implemented using PyTorch 1.11.0 and trained on an NVIDIA GeForce RTX 3090 GPU. 

For the hardware performance assessment, the proposed PatchINR accelerator is implemented and deployed on a high-performance Xilinx Virtex UltraScale+ HBM VCU128 FPGA platform. The hardware design is synthesized, placed, and routed using Xilinx Vivado 2019.2 design suite, targeting an operating frequency of 200 MHz. To validate the flexibility and efficiency of our architecture, the hardware evaluation encompasses two distinct computational modes: standard 32-bit floating-point (FP32) and highly efficient 8-bit integer (INT8) quantization. 
\subsection{Experiment Results}
\begin{figure}[!t]
\centering
\includegraphics[width = 1.0\columnwidth]{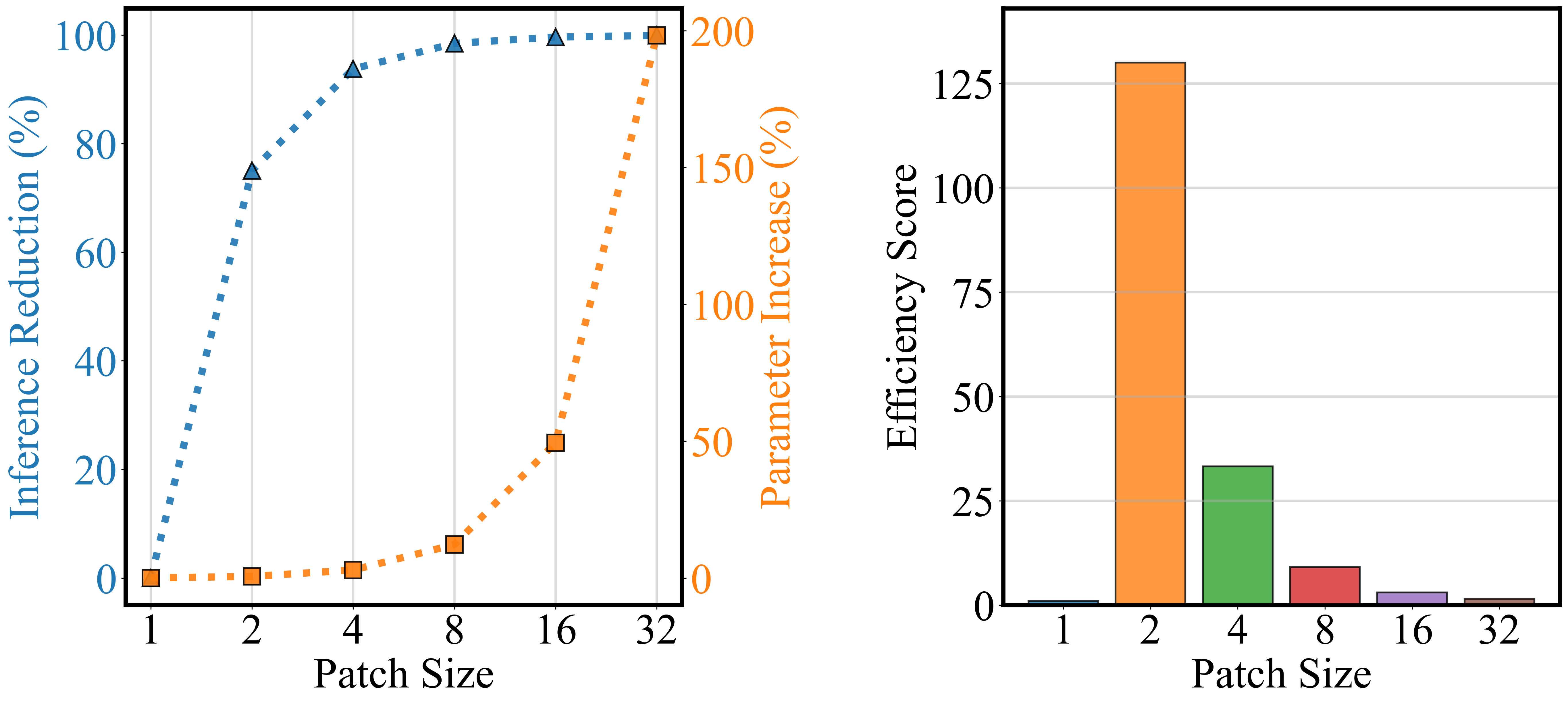}
\vspace{-0.3cm}
\caption{Patch optimization efficiency analysis. (Left) Inference reduction vs. parameter increase trade-off. (Right) Efficiency scores across patch sizes.}
\label{fig:patch_experiment}
\end{figure}

To quantitatively evaluate patch-based optimization in INR networks, we propose an efficiency metric balancing inference reduction benefits against parameter overhead costs. For patch size N, the efficiency score E is:
\begin{equation}
E(N) = \frac{IR(N)}{1 + PI(N)}
\end{equation}
where $IR(N)$ and $PI(N)$ represent the inference reduction and parameter increase percentages relative to patch size 1. Fig.~\ref{fig:patch_experiment} shows that patch size $N=2$ achieves optimal efficiency with significant inference reduction and negligible parameter overhead, while larger patches exhibit diminishing returns due to exponential parameter growth.

Table~\ref{tab:result} compares the performance of image reconstruction for different patch sizes (\(N\)) across three different INR architectures. Increasing the patch size leads to consistent improvements in both PSNR and SSIM for all models, demonstrating the generalization ability of our method. From the patch size perspective, it is evident that increasing the patch size brings performance gains, but the growth is not linearly increasing. When the patch size reaches 16, the PSNR only improves around 4 to 5 dB, but further doubling the patch size brings more than 10 dB improvement.
\begin{table}[!t]
\centering
\caption{Comparison of INR architectures (SIREN~\cite{siren}, WIRE~\cite{wire}, and FINER~\cite{finer}) on Kodak image reconstruction tasks across varying patch sizes \( N \).}
\label{tab:result}
\renewcommand{\arraystretch}{1.3}
\resizebox{\linewidth}{!}{
\begin{tabular}{lcccccc}
\toprule
\multirow{2}{*}{\textbf{Patch Size}} 
& \multicolumn{2}{c}{\textbf{WIRE~\cite{wire}}} 
& \multicolumn{2}{c}{\textbf{FINER~\cite{finer}}} 
& \multicolumn{2}{c}{\textbf{SIREN~\cite{siren}}} \\ \cmidrule(lr){2-3} \cmidrule(lr){4-5} \cmidrule(lr){6-7}
& \textbf{PSNR (dB)} & \textbf{SSIM} 
& \textbf{PSNR (dB)} & \textbf{SSIM} 
& \textbf{PSNR (dB)} & \textbf{SSIM} \\ \hline
$N=1$                     & 33.36       & 0.9380        & 33.28        & 0.9707        & 32.80        & 0.9912        \\
\rowcolor{blue!10} $N=2$                     & 34.97       & 0.9549        & 34.54        & 0.9774        & 34.37        & 0.9944        \\
$N=4$                     & 35.26       & 0.9326        & 35.53        & 0.9183        & 34.52        & 0.8786        \\
$N=8$                     & 35.46       & 0.9624        & 36.54        & 0.9819        & 36.01        & 0.9953        \\
$N=16$                    & 37.96       & 0.9750        & 38.40        & 0.9825        & 37.52        & 0.9921        \\
$N=32$                    & 44.05       & 0.9893        & 44.73        & 0.9877        & 41.86        & 0.9884        \\
\bottomrule
\end{tabular}
}
\end{table}
\begin{figure}[!t]
\centering
\includegraphics[width = 1.0\columnwidth]{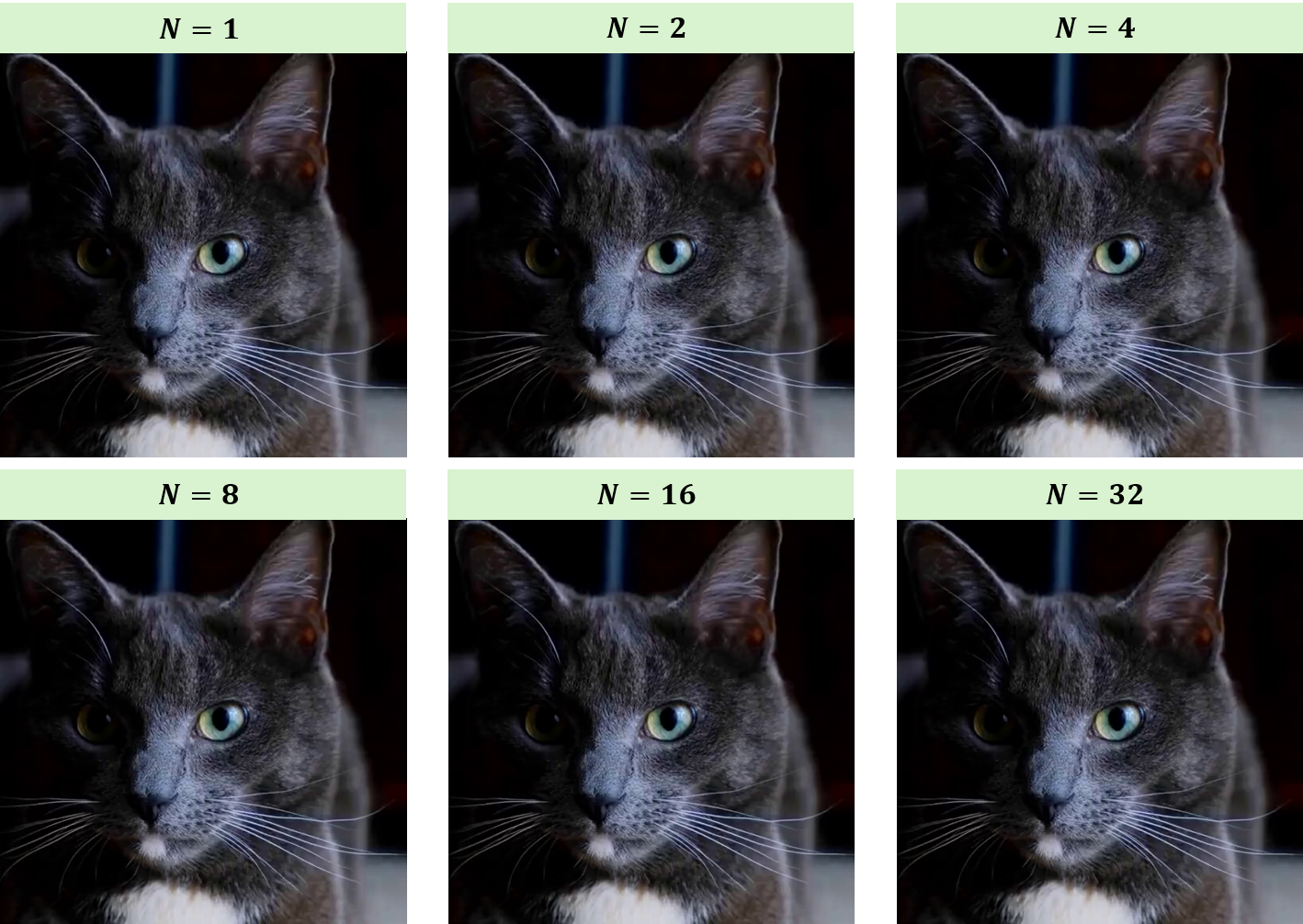}
\vspace{-0.3cm}
\caption{Comparison of video reconstruction performance across varying patch sizes ($N$).}
\label{fig:visual_patch}
\vspace{-0.3cm}
\end{figure}

In addition to image tasks, we further validate our method on the video modality and provide visual examples in Fig.~\ref{fig:visual_patch} across various patch sizes \(N\). Video can be viewed as a series of images connected along the time axis, and our patch-based method can be applied to each frame. Besides the superior reconstruction quality, our method significantly reduces the training time for video, a critical challenge for video INR. Previous approaches mainly address this limitation from a frame perspective, that is, taking the frame index as input and directly outputting the whole frame, which sacrifices the most appealing resolution-independent property of INR.

Table~\ref{table:performance_summary} highlights the key architectural advantages of our FPGA-based accelerator through its focused resource allocation. The high concentration of DSPs and power in the MAC array reflects a co-design strategy that directly maps hardware to the core computational workload, maximizing efficiency. Our fully PL-based, custom datapath eliminates the overhead of fixed-function units, providing greater optimization flexibility compared to hybrid AI Engine architectures. Furthermore, the minimal control logic and resource-efficient implementation of sinusoidal activation demonstrate a comprehensive optimization across the entire computational pipeline.
\begin{table}[!t]
\centering
\caption{Performance Summary and Hardware Resource Breakdown of the Proposed FP32 INR Accelerator on VCU128 at 200 MHz}
\label{table:performance_summary}
\renewcommand{\arraystretch}{1.3}
\resizebox{\linewidth}{!}{
\begin{tabular}{lcccccc}
\toprule
\textbf{Component} & \multicolumn{5}{c}{\textbf{Resource Utilization}} & \textbf{Power} \\
\cmidrule(lr){2-6}
\textbf{(Five layers)} & \textbf{LUT} & \textbf{FF} & \textbf{DSP} & \textbf{LUTRAM} & \textbf{BRAM} & \textbf{(mW)} \\
\midrule
MAC Array & 244,759 & 442,268 & 5,130 & 14,394 & 0 & 6,350  \\
Top Ctrl. & 60 & 55 & 0 & 0 & 0 & 10  \\
BRAM & 0 & 0 & 0 & 0 & 231 & 615  \\
Nonl. Act. & 5,032 & 388 & 8 & 0 & 0 & 112  \\
Others & 525 & 99,015 & 0 & 396 & 0 & 455  \\
\rowcolor{blue!10} Total & 250,376 & 541,726 & 5,138 & 14,790 & 231 & 7,542  \\
\bottomrule
\end{tabular}
}
\end{table}
\begin{figure}[!t]
\centering
\includegraphics[width = 1.0\columnwidth]{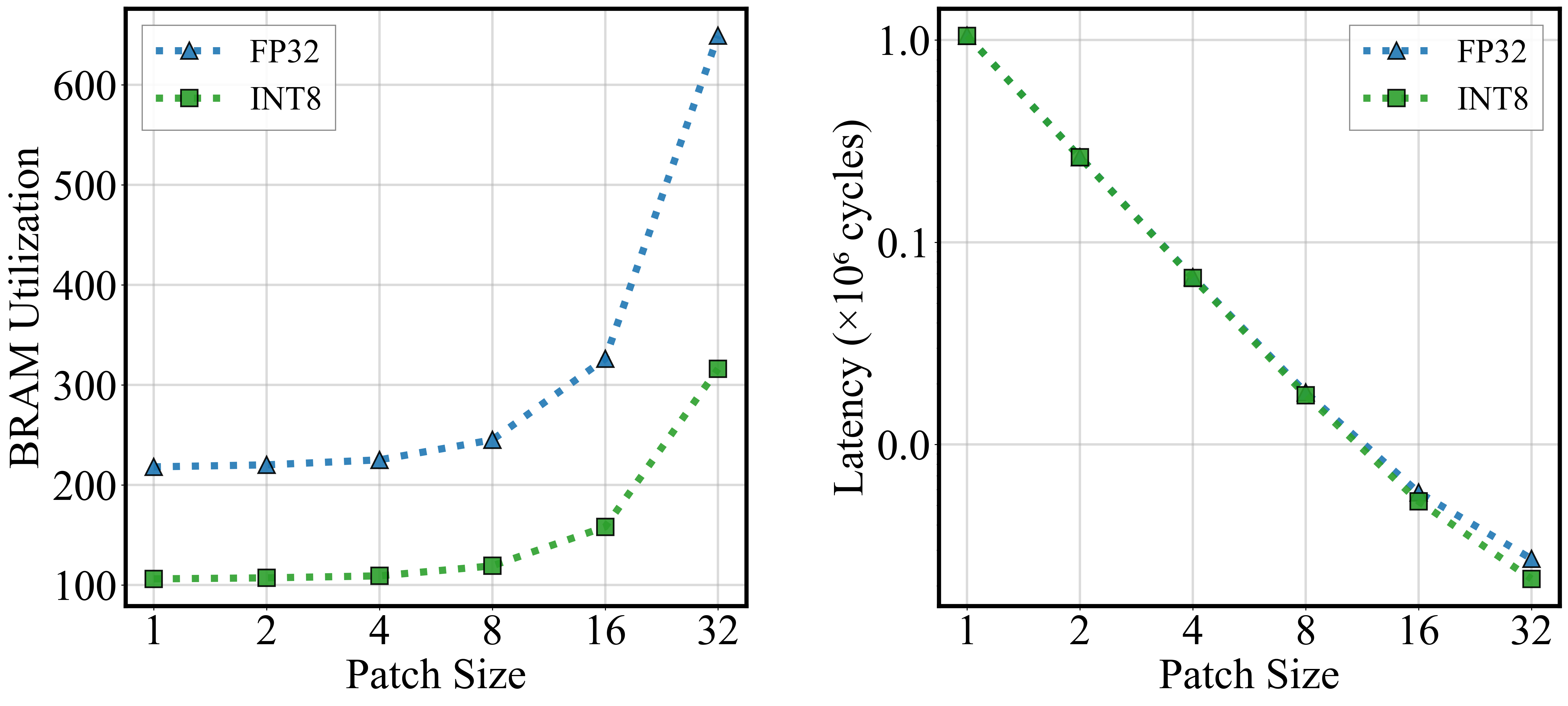}
\vspace{-0.3cm}
\caption{FPGA Performance Analysis. (Left) BRAM utilization and (Right) inference latency for FP32 and INT8 precision modes across different patch sizes.}
\label{fig:fpga_analysis}
\vspace{-0.3cm}
\end{figure}

Fig.~\ref{fig:fpga_analysis} presents the BRAM utilization and inference latency across different patch sizes. For BRAM utilization, the INT8 mode consistently shows lower resource consumption than the FP32 mode, maintaining relatively stable utilization for small patches but exhibiting exponential growth beyond patch size 16. The latency analysis demonstrates a critical transition in performance bottleneck: at small patch sizes, total latency is primarily determined by the number of inference operations, exhibiting significant reduction with increased patch size. As patch size exceeds 16, the hardware pipeline latency becomes the dominant factor, leading to diminishing returns in latency reduction. These findings suggest an optimal patch size selection should balance between latency reduction and resource utilization, with small patches achieving the highest efficiency due to significant latency reduction while maintaining minimal resource overhead.
\section{Conclusion}
\label{sec:conclusion}
In this work, we introduced a patch-based INR framework that predicts entire $n \times n$ pixel patches in a single forward pass, reducing computational overhead while maintaining resolution independence and continuous reconstruction capabilities. To validate the effectiveness of our approach, we propose a hardware acceleration architecture on the FPGA platform for the SIREN model, which features a configurable pipeline and supports dual-precision computation. Our fixed-size patch output structure achieves hardware efficiency comparable to pixel-wise INRs with superior reconstruction quality at larger patch sizes. Our method achieves an optimal trade-off between inference efficiency and parameter count, where a $2 \times 2$ patch size reduces inference latency by 75\% at 34.97 dB PSNR with only 0.6\% parameter overhead.

\section*{Acknowledgment}

This research is supported in part by the Theme-based Research Scheme (TRS) project T45-701/22-R and GRF Project 17203224 of the Research Grants Council (RGC), Hong Kong SAR, and in part by the AVNET-HKU Emerging Microelectronics \& Ubiquitous Systems (EMUS) Lab.

\bibliographystyle{ieeetr}
\bibliography{bibliography}
\end{document}